\documentclass[10pt]{wlscirep}
\usepackage[T1]{fontenc}
\usepackage[utf8]{inputenc}
\usepackage{babel}
\usepackage{csquotes}

\usepackage[numbers,sort&compress]{natbib}

\usepackage{bm}
\usepackage{physics}
\usepackage{mathtools}
\usepackage{tcolorbox}
\usepackage{tikz}
\usepackage[siunitx]{circuitikz}
\usetikzlibrary{positioning}
\usepackage{subcaption}
\usepackage{todonotes}
\usepackage[percent]{overpic} 
\renewcommand{\vb}[2]{#2}
\usepackage[edges]{forest}

\usepackage{tikz} 
\usepackage{pgfplots}

\usetikzlibrary{shapes.geometric,decorations,backgrounds}
\usetikzlibrary{arrows.meta}
\usetikzlibrary{chains,arrows}
\usetikzlibrary{automata, positioning, arrows}
\usepackage[outline]{contour} 
\usetikzlibrary{calc}
\usetikzlibrary{angles,quotes} 
\contourlength{1.2pt}
\usetikzlibrary{decorations.pathmorphing}
\usetikzlibrary{fit}

\tikzset{snake it/.style={decorate, decoration=snake}}

\tikzset{
        >=latex, 
        node distance=2cm, 
        }
        
\definecolor{myred}{RGB}{169,44,31}
\definecolor{mygreen}{RGB}{100,200,55}
\definecolor{mypink}{RGB}{255,200,200}
\definecolor{myblue}{RGB}{58,132,186}
\definecolor{myyellow}{RGB}{243,192,105}
\definecolor{mygray}{RGB}{242,242,242}

\definecolor{mylightred}{RGB}{255,150,150}
\definecolor{mylightgreen}{RGB}{200,200,255}
\definecolor{mylightblue}{RGB}{200,200,255}
\definecolor{mylightyellow}{RGB}{243,192,105}
\colorlet{xcol}{blue!70!black}
\colorlet{vcol}{green!60!black}
\colorlet{mydarkred}{myred!70!black}
\colorlet{mydarkblue}{myblue!60!black}
\colorlet{mydarkgreen}{mygreen!60!black}
\colorlet{acol}{red!50!blue!80!black!80}
\tikzstyle{CM}=[red!40!black,fill=red!80!black!80]
\tikzstyle{xline}=[xcol,thick,smooth]
\tikzstyle{mass}=[line width=0.6,red!30!black,fill=red!40!black!10,rounded corners=1,
                  top color=red!40!black!20,bottom color=red!40!black!10,shading angle=20]
\tikzstyle{faded mass}=[dashed,line width=0.1,red!30!black!40,fill=red!40!black!10,rounded corners=1,
                        top color=red!40!black!10,bottom color=red!40!black!10,shading angle=20]
\tikzstyle{rope}=[brown!70!black,very thick,line cap=round]

\tikzstyle{force}=[->,myred,very thick,line cap=round]
\tikzstyle{velocity}=[->,vcol,very thick,line cap=round]
\tikzstyle{Fproj}=[force,myred!40]
\tikzstyle{myarr}=[-{Latex[length=3,width=2]},thin]

\title{Neuromorphic Intelligence}
\author[1,*]{Marcel van Gerven}
\affil[1]{Department of Machine Learning and Neural Computing, Donders Institute for Brain, Cognition and Behaviour, Radboud University, Nijmegen, the Netherlands}

\affil[*]{e-mail: marcel.vangerven@donders.ru.nl}

\begin{abstract}
Neuromorphic computing seeks to replicate the remarkable efficiency, flexibility, and adaptability of the human brain in artificial systems. Unlike conventional digital approaches, which suffer from the Von Neumann bottleneck and depend on massive computational and energy resources, neuromorphic systems exploit brain-inspired principles of computation to achieve orders of magnitude greater energy efficiency. By drawing on insights from a wide range of disciplines -- including artificial intelligence, physics, chemistry, biology, neuroscience, cognitive science and materials science -- neuromorphic computing promises to deliver intelligent systems that are sustainable, transparent, and widely accessible. A central challenge, however, is to identify a unifying theoretical framework capable of bridging these diverse disciplines. We argue that dynamical systems theory provides such a foundation. Rooted in differential calculus, it offers a principled language for modeling inference, learning, and control in both natural and artificial substrates. Within this framework, noise can be harnessed as a resource for learning, while differential genetic programming enables the discovery of dynamical systems that implement adaptive behaviors. Embracing this perspective paves the way toward emergent neuromorphic intelligence, where intelligent behavior arises from the dynamics of physical substrates, advancing both the science and sustainability of AI.\\


\end{abstract}

\pgfplotsset{compat=1.18}
\begin{document}
\flushbottom
\maketitle



\section{Introduction}

Artificial intelligence (AI) -- defined as the science and engineering of creating intelligent machines~\cite{McCarthy2007What} -- has experienced unprecedented advancements over the past decade. This remarkable progress stems largely from the deep learning revolution~\cite{LeCun2015Deep}. That is, the ability to train deep neural networks (DNNs) at scale using the backpropagation algorithm~\cite{Werbos1974Regression} in conjunction with abundant access to data and compute, has propelled AI from laboratory settings into mainstream society. This transformation is evidenced by significant scientific breakthroughs across multiple disciplines~\cite{Abramson2024Accurate,Trinh2024Solving,Kolmus2022Fast} and the emergence of applications addressing diverse societal challenges~\cite{Tuia2022Perspectives,Rolnick2022Tackling,Dingemans2023PhenoScore}.
Perhaps most striking is the development of generative AI models with unprecedented reasoning capabilities~\cite{OpenAI2024GPT4,Touvron2023Llama,Gu2024Mamba}.
 Motivated by neural scaling laws, these breakthroughs have prompted some to suggest that achieving human-level artificial general intelligence only requires more data, memory, and compute to train end-to-end models at scale~\cite{SuttonRich2019Bitter,Kaplan2020Scaling}. 


Despite the significant advancements and potential of deep learning to drive scientific breakthroughs and address societal challenges, the current approach to AI also comes with considerable drawbacks. Training and using DNNs at scale consumes vast amounts of natural resources and results in a substantial carbon footprint, which negatively impacts the environment~\cite{Strubell2019Energy,Schwartz2020Green}. On its current exponential trajectory, the cost of compute is projected to approach the world's total energy production by 2050 with AI as one of its major contributors~\cite{AngEtAl.2021Decadal,Mehonic2022Braininspired}. The problem with the scale-is-all-you-need approach to AI is, of course, that it is unsustainable in a resource-bounded world~\cite{Steffen2015Planetary}.\footnote{ This issue is receiving increasing attention~\cite{Elsworth2025Measuring}. At the same time, AI may provide breakthroughs that help address global energy needs~\cite{Degrave2022Magnetic,Pyzer-Knapp2025Foundation,Meselhy2025Review}.}
The resource-intensive nature of this approach further implies that it is predominantly accessible to big tech companies, whose policies can have adverse societal and geopolitical consequences~\cite{Crawford2021Atlas}. 
Another challenge associated with the trend of developing increasingly complex and therefore opaque deep learning models is that it makes it progressively difficult to understand their functioning~\cite{Ras2022Explainable}. 
Finally,
the present emphasis on scale and application-driven progress leaves limited space for addressing the fundamental question which originally inspired the establishment of AI as a scientific discipline~\cite{LighthillJames1973Artificial,Lake2017Building,Hassabis2017Neuroscienceinspired,VanGerven2017Computational}, that is, how does mind emerge from matter? Indeed, despite the remarkable advances in AI, the human brain remains superior in terms of its reasoning capabilities and adaptability in real-world settings~\cite{Marcus2020Next}.



These considerations raise the question of whether there is an alternative approach to AI that may be more sustainable and aligned with the ambition to understand the nature of intelligence. 
Neuromorphic computing, which mimics the neural information processing mechanisms of the human brain, offers a route forward~\cite{Mead1990Neuromorphic,Modha2011Cognitive}. 
The human brain is estimated to achieve approximately one exaFLOPS ($10^{18}$ floating-point operations per second) while consuming only 20 W of power~\cite{Hassabis2017Neuroscienceinspired,Marcus2019Rebooting}, equivalent to powering one light bulb.\footnote{In contrast, the El Capitan supercomputer operates at 1.7 exaFLOPS while consuming 30 MW of power, which is sufficient to power a small town.}  
Neuromorphic computing, or brain-inspired computing, aims to develop systems as flexible and efficient as the human brain by embracing its structure and function 
and replicating this in physical devices~\cite{Mead1990Neuromorphic,Indiveri2011Neuromorphic,Neftci2013Synthesizing,Chicca2020Recipe}.
Hence, mimicking natural computing in artificial systems has the potential to significantly reduce energy consumption and environmental impact, as neuromorphic systems are designed to be highly efficient in terms of power usage~\cite{Davies2018Loihi}. Moreover, it provides a bridge between natural and artificial intelligence~\cite{Indiveri2025Neuromorphic}.

While neuromorphic computing holds great promise as an attractive alternative to traditional deep learning approaches, we have not yet fully realized its potential~\cite{Mehonic2022Braininspired,Kudithipudi2025Neuromorphic}. The development and deployment of neuromorphic systems require a concerted effort across multiple disciplines, including the natural sciences, materials science, artificial intelligence and systems engineering~\cite{AngEtAl.2021Decadal,Mentink2025Neuromorphic,Pedersen2024Neuromorphic}. 
The natural sciences provide insights into the mechanisms employed by nature from different perspectives, including physics~\cite{Markovic2020Physics,Chen2020Classification},  chemistry~\cite{VanDeBurgt2017Nonvolatile}, biology~\cite{Smirnova2024Biocomputing}, neuroscience~\cite{Hassabis2017Neuroscienceinspired,Zenke2021Visualizing,Kagan2022Vitro,Zador2023Catalyzing,Sadeh2025Emergence} and cognitive science~\cite{Gelder1998Dynamical}. Materials science is crucial for designing and fabricating specialized hardware that emulates biological information processing, ensuring energy efficiency and high performance \cite{Davies2018Loihi,Kaspar2021Rise}. Artificial intelligence -- in particular the subfield of machine learning -- provides algorithms and models that leverage neuromorphic hardware's unique capabilities~\cite{Schuman2022Opportunities,Jaeger2023Formal}. Finally, systems engineering integrate these components into scalable architectures, managing information flow effectively in real-world applications~\cite{Muir2025Road,Christensen20222022}. 

To realize the full potential of neuromorphic computing, it is essential to establish a comprehensive program that encourages collaboration among this diversity of disciplines. 
The key to achieving this multidisciplinary integration relies on the adoption of a common theoretical framework that unites these disciplines. Dynamical systems theory (DST) is the natural candidate for such a framework~\cite{Strogatz1994Nonlinear}. 
Dynamical systems theory adopts differential calculus -- which has proven extremely successful in other branches of science~\cite{Newton1687Philosophiae,Einstein1905Uber,Langevin1997Theorie} -- to model the mechanisms underlying (natural and artificial) intelligence. As such, it views AI as a discipline rooted in the natural sciences. 
This dynamical systems approach to the development of intelligent systems dates back to the early days of AI, which has its roots in the cybernetics movement~\cite{Kline2011Cybernetics}. Already in 1947, the cyberneticist Ross Ashby, described self-organization of complex systems in terms of dynamical systems that automatically evolve towards an attractor state in a basin of surrounding states~\cite{Ashby1947Principles}. The dynamical systems approach to AI has been advocated several times before~\cite{Amari1977Dynamics,Cohen1983Absolute,Pearlmutter1990Dynamic,Beer1992Evolving,Beer1995Dynamical,Beer1995Dynamics,Beer1997Dynamics} but, given the dominance of digital computing based on the von Neumann paradigm, 
has not been widely adopted by the AI community to guide the development of intelligent systems. In contrast, in the neuromorphic computing community, dynamical systems theory is the language of choice given the objective to mimic natural intelligence in (noisy, analog) physical hardware~\cite{Mead1990Neuromorphic,Jaeger2023Formal}. 

By embracing dynamical systems theory, we can bridge between algorithms and models motivated by the natural sciences and implementations on neuromorphic substrates in order to realize AI systems that are more aligned with natural intelligence.   Adopting this common language also allows for a fruitful exchange with several other scientific disciplines that already use DST as their theoretical foundation~\cite{Murray2002Mathematical,Lorenz1993Nonlinear,Castellano2009Statistical, Izhikevich2018Dynamical, Breakspear2017Dynamic, Thelen1994Dynamic, Spivey2008Continuity,Krakauer2024Complex}. 
As will be shown, dynamical systems can be set up in such a way that inference, learning and control ensues purely by following the system's equations of motion. By mapping these equations directly onto physical substrates we remove the distinction between software and hardware, driving the development of neuromorphic systems. Finally, I argue that intelligent neuromorphic systems can be automatically discovered using evolutionary algorithms that are able to search through the space of (constrained) dynamical systems. This provides a truly end-to-end approach to realizing (bottom-up, emergent) {\em neuromorphic intelligence}.

\section{The dynamical systems approach to neuromorphic intelligence}


Neuromorphic computing aims to develop systems as flexible and efficient as the human brain by embracing its analog or mixed-signal  (analog-digital) nature and replicating this in physical devices~\cite{Mead1990Neuromorphic,Mahowald1991Silicona,Indiveri2011Neuromorphic,Chicca2020Recipe}. The question remains how to construct neuromorphic systems such that they replicate principles and mechanisms underlying natural intelligence. We will approach this question from a dynamical systems perspective. 
To introduce the dynamical systems approach to neuromorphic intelligence, we will find it convenient to adopt Marr's proposal that complex information processing systems should be considered at three different levels of analysis~\cite{Marr1976Understanding,VanGerven2017Computational,Guo2021Marrs}, that is, the computational, algorithmic and implementational level. 
Figure~\ref{fig:marr-neuromorphic}a depicts these three levels of analysis from the perspective of neuromorphic computing.


 \newcommand{\circuit}{%
\scalebox{0.3}{
  \begin{tikzpicture}
\begin{circuitikz}[american]
  \draw (-3,0) to[C, l=$C_m$] (-3,4);

  \draw (-3,4) -- (-1,4)
        to[R, l=$R_L$] (-1,2)
        to[battery1, l_=$E_L$] (-1,0)
        -- (-3,0);

  \draw (-1,4) -- (1,4)
        to[vR, l=$R_K$] (1,2)
        to[battery1, l_=$E_K$] (1,0)
        -- (-1,0);

  \draw (1,4) -- (3,4)
        to[vR, l=$R_{Na}$] (3,2)
        to[battery1, l_=$E_{Na}$] (3,0)
        -- (1,0);

  \node[ground] (ground) at (0,0) {};

  \draw (0,4.5) to [short, *-] (0,4);
\end{circuitikz}
  \end{tikzpicture}%
}}

\begin{figure}[!ht]
\centering
\begin{subfigure}[t]{0.32\textwidth}
\centering
   \begin{picture}(0,0)
      \put(5,170){\makebox(0,0)[l]{(a)}} 
    \end{picture}
\scalebox{1.1}{
\begin{scriptsize}
\begin{tikzpicture}
    [>=Stealth, node distance=10pt, minimum size=0.3cm, inner sep=1pt]


\node (a1) {{\bf Computational level}};
\node (b1) [below of=a1] {$a^* = \underset{a \in \mathcal{A}}{\arg\max}\, J(a)$};
\node[fit=(a1)(b1), draw, inner sep=3pt, minimum width=3.5cm, rounded corners] (block1) {};

\node[below=20pt of block1] (a2) {{\bf Algorithmic level}};
\node (b2) [below of=a2] {$\dd z = f_z(x) \dd v_z$};
\node[fit=(a2)(b2), draw, inner sep=3pt, minimum width=3.5cm, rounded corners] (block2) {};



\node[below=20pt of block2] (a3) {{\bf Implementational level}};
\node (b3) [baseline=(current bounding box.north)] at ($(a3) + (-1.15,0.4)$) {\circuit};

\node (n3) at (0,-4.8) {};
\node[fit=(a3)(n3), draw, inner sep=5pt, minimum width=3.5cm, rounded corners] (block3) {};

\draw[->] ($ (block1.south) + (-0.5cm,0) $) -- node[left, xshift=-1mm]{justifies} ($ (block2.north) + (-0.5cm,0) $);
\draw[->] ($ (block2.north) + (0.5cm,0) $) -- node[right, xshift=1mm]{specifies} 
($ (block1.south) + (0.5cm,0) $);

\draw[->] ($ (block2.south) + (-0.5cm,0) $) -- node[left, xshift=-1mm]{constrains} ($ (block3.north) + (-0.5cm,0) $);
\draw[->] ($ (block3.north) + (0.5cm,0) $) -- node[right, xshift=1mm]{realizes} ($ (block2.south) + (0.5cm,0) $);


\end{tikzpicture}
\end{scriptsize}
}
\end{subfigure}
\begin{subfigure}[t]{0.67\textwidth}
\centering
   \begin{picture}(0,0)
      \put(-10,170){\makebox(0,0)[l]{(b)}} 
    \end{picture}
\scalebox{1.2}{
\begin{scriptsize}
\begin{tikzpicture}
    [>=Stealth, node distance=2cm, minimum size=0.3cm, inner sep=1pt]
    
    
    \node (z0) [draw=black, circle, fill=mylightblue] at (0,-1.5) {$z_0$};
    \node (z1) [draw=black, circle, fill=mylightblue, right of=z0] {$z_1$};
    \node (z2) [draw=black, circle, fill=mylightblue, right of=z1] {$z_2$};
    \node (z3) [draw=black, circle, fill=mylightblue, right of=z2] {$z_3$};
    \node (zdots) [right= 0.5 of z3] {$\cdots$};

    \scoped[on background layer]
    \draw [rounded corners, fill=mygray,draw] (z0)+(-1,0.5) rectangle ([xshift=1cm,yshift=-0.5cm]zdots) {};

    \draw[->, >=latex] (z0) -- (z1);
    \draw[->, >=latex] (z1) -- (z2);
    \draw[->, >=latex] (z2) -- (z3);   

 \node (u0) [draw=black,  fill=mylightblue] at (0.0,-0.6) {$u_0$};
    \node (u1) [draw=black, minimum size=0.3cm, fill=mylightblue, right of=u0] {$u_1$};
    \node (u2) [draw=black,  fill=mylightblue, right of=u1] {$u_2$};
    \node (u3) [draw=black,  fill=mylightblue, right of=u2] {$u_3$};

    \draw[->, >=latex] (z0) -- (u0);
    \draw[->, >=latex] (z1) -- (u1);
    \draw[->, >=latex] (z2) -- (u2);
    \draw[->, >=latex] (z3) -- (u3);
    
    \node (s0) [draw=black, circle, fill=mylightblue] at (0,1.5) {$s_0$};
    \node (s1) [draw=black, circle, fill=mylightblue, right of=s0] {$s_1$};
    \node (s2) [draw=black, circle, fill=mylightblue, right of=s1] {$s_2$};
    \node (s3) [draw=black, circle, fill=mylightblue, right of=s2] {$s_3$};
    \node (sdots) [right= 0.5 of s3] {$\cdots$};

    \scoped[on background layer]
    \draw[rounded corners, fill=mygray,draw] (s0)+(-1,0.5) rectangle ([xshift=1cm,yshift=-0.5cm]sdots) {};

    \draw[->, >=latex] (s0) -- (s1);
    \draw[->, >=latex] (s1) -- (s2);
    \draw[->, >=latex] (s2) -- (s3);   

 \node (y0) [draw=black, fill=mylightblue] at (0,0.6) {$y_0$};
    \node (y1) [draw=black, fill=mylightblue, right of=y0] {$y_1$};
    \node (y2) [draw=black, fill=mylightblue, right of=y1] {$y_2$};
    \node (y3) [draw=black,  fill=mylightblue, right of=y2] {$y_3$};

    \draw[->, >=latex] (s0) -- (y0);
    \draw[->, >=latex] (s1) -- (y1);
    \draw[->, >=latex] (s2) -- (y2);
    \draw[->, >=latex] (s3) -- (y3);

    \draw[->, >=latex] (u0) -- (s1);
    \draw[->, >=latex] (u1) -- (s2);
    \draw[->, >=latex] (u2) -- (s3);

    \draw[->, >=latex] (y0) -- (z1);
    \draw[->, >=latex] (y1) -- (z2);
    \draw[->, >=latex] (y2) -- (z3);

\node (r0) [draw=black, diamond, inner sep=.5pt, fill=mylightblue] at (0.0,0.0) {$r_0$};
    \node (r1) [draw=black, diamond, inner sep=0.5pt,   fill=mylightblue, right of=r0] {$r_1$};
    \node (r2) [draw=black, diamond, inner sep=.5pt, fill=mylightblue, right of=r1] {$r_2$};
    \node (r3) [draw=black, diamond, inner sep=.5pt, fill=mylightblue, right of=r2] {$r_3$};

  \draw[->, >=latex] (z0) to[out=135, in=225] (r0);
  \draw[->, >=latex] (z1) to[out=135, in=225] (r1);
  \draw[->, >=latex] (z2) to[out=135, in=225] (r2);
  \draw[->, >=latex] (z3) to[out=135, in=225] (r3);

  \draw[->, >=latex] (s0) to[out=225, in=135] (r0);
  \draw[->, >=latex] (s1) to[out=225, in=135] (r1);
  \draw[->, >=latex] (s2) to[out=225, in=135] (r2);
  \draw[->, >=latex] (s3) to[out=225, in=135] (r3);

\node at ($(z0)!0.5!(zdots)$) [yshift=-0.75cm,font=\sffamily] {\scriptsize{Agent}}; 
  \node at ($(s0)!0.5!(sdots)$) [yshift=0.75cm,font=\sffamily] {\scriptsize{Environment}};

\end{tikzpicture}
\end{scriptsize}
}
\end{subfigure}
\caption{The dynamical systems approach to neuromorphic intelligence. (a) Marr's three levels of analysis from the perspective of neuromorphic computing. The computational level makes explicit what the computational problem is. The algorithmic level defines how the computational problem is solved. The implementational level makes explicit how this solution is realised in the neuromorphic substrate. (b) An agent and its environment can be modelled as a coupled dynamical system with state $x = (z, s)$. 
Interactions between the agent and the environment are mediated by observations $y$ 
and controls $u$. 
The agent's objective is quantified in terms of instantaneous rewards $r$, which may depend on the state of both the agent and the environment. State variables are denoted by circles, system inputs by squares and rewards by diamonds. The continuous time dynamics are recovered as the time step $\Delta t$ between transitions goes to zero.}
\label{fig:marr-neuromorphic}
\end{figure}
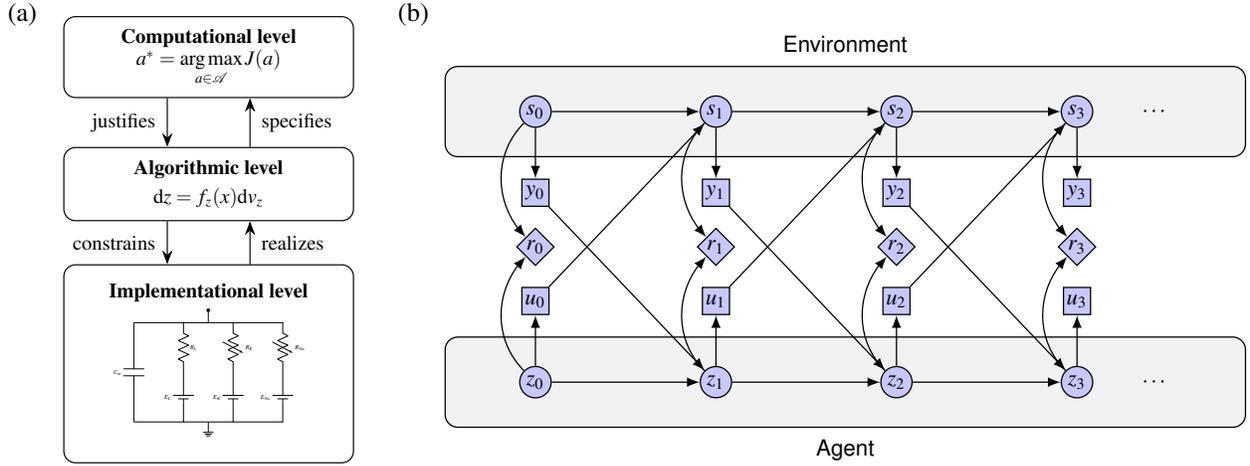

\subsection{The computational level}
At the computational level of analysis, the goal is to understand the computational problem that is solved by a natural agent and to design artificial agents that solve the same problem. 
Let us formulate this problem in abstract terms as the ability of an agent to adapt its behavior in order to optimally achieve its objective~\cite{VanGerven2017Computational}. That is, an agent $a$ is a (rational, intelligent) entity which acts in a {\em seemingly} teleological (goal-directed) manner in some environment $e$~\cite{Russell2021Artificial}.

To formalize such goal-directness, we assume the  existence of some objective function $J$ which the agent strives to optimize~\cite{Marblestone2016Integration,Richards2019Deep}. 
For concreteness, we  assume that the objective function of an agent $a$ operating in environment $e$ 
takes the form of an expected discounted return (cumulative reward), given by 
\begin{equation}
J_e(a) 
= \mathbb{E}\left[\int_{0}^{\infty} \alpha(t) r(t) \dd t \mid A = a, E = e\right]
\label{eq:objective}
\end{equation}
where  
$\alpha(t) = \exp(-\lambda t)$ is an (exponential) discounting function, and the instantaneous reward $r(t)$ is an implicit function of the state of the agent and/or environment.
The expectation operator expresses that the discounted return is a random variable due to the stochastic nature of the coupled dynamics of the agent and the environment. We also refer to such irreducible uncertainty as aleatoric uncertainty.  
%
We proceed by defining the {\em intelligence} of an agent as
\begin{equation}
J(a) 
= \mathbb{E}_{e \sim p_E}  \left[ J_e(a) \right] 
\label{eq:intelligence}
\end{equation}
where the expectation is taken over the environmental distribution $p_E$. That is, an agent's intelligence is defined as the expected ability of an agent to behave optimally in any environment. 
It follows that, at the computational level of analysis, the goal is to find the optimal agent
\begin{equation}
    a^* = \arg\max_{a \in \mathcal{A}} J(a) 
\end{equation}
where $\mathcal{A}$ denotes the set of all agents. 

\subsection{The algorithmic level}

The algorithmic level of analysis asks how an intelligent system behaves the way it does. That is, what are the mechanisms that allow the agent to achieve its computational-level goal? 
The dynamical systems approach assumes that both the agent and the environment are structured physical processes that unfold over time, subject to external influences. 




Formally, a dynamical system  is a rule for time evolution on a state space, which can be described in terms of (a system of) differential equations. 
Motivated by Carver Mead's quote that ``drift and diffusion are the stuff that biology and neuromorphic systems are made of''~\cite{Mead1990Neuromorphic}, we will assume that neuromorphic systems can be suitably expressed in terms of a (vector-valued) stochastic differential equation (SDE). Let $x(t) \in \mathbb{R}^d$ denote a state vector. Given an initial state $x(0) = x_0$, the time evolution of $x(t)$ is described by an SDE
\begin{equation}
\dd {x} = \mu(\vb*{x})  \dd{t} + \sigma(\vb*{x})\dd \xi
\label{eq:SDE}
\end{equation}
where the drift function $\mu \colon \mathbb{R}^d \to \mathbb{R}^d$ and diffusion function $\sigma  \colon \mathbb{R}^d \to \mathbb{R}^{d \times n}$ represent the deterministic and stochastic components of the dynamics, respectively, and $\xi(t)$ is the realization of some noise process. 
Note that the time indices for $x(t)$ and $\xi(t)$ are omitted in Eq.~\eqref{eq:SDE} for ease of exposition. To further simplify notation, we will represent the SDE~\eqref{eq:SDE} in more compact form as 
$
\dd x = f(x) \dd v
$, where $f \colon \mathbb{R}^{d}  \to \mathbb{R}^{d \times n}$ is the state equation and $v(t) \in \mathbb{R}^{n}$ are the driving paths, typically representing time and/or uncontrollable random forces. We may write Eq.~\eqref{eq:SDE} in this more compact form by defining $f(x) = [\mu(x), \sigma(x)]$ and $v= (t, \xi)$, where we use
 $(v_1,\ldots, v_n)$ to denote vertical stacking of (the elements in) $v_1,\ldots,v_n$.

\begin{tcolorbox}[float=t, title=Box 1: Controlling a stochastic particle, colback=white, colframe=blue, colbacktitle=lightgray, coltitle=black, fonttitle=\bfseries]
\label{box:box1}

Let us consider positional control of a stochastic particle as a minimal working example. We can represent this problem using coupled stochastic differential equations.
Consider a particle whose state is given by $s = (s_1, s_2) = (p, \dot{p})$ with $p$ the particle position and $\dot{p}$ the particle velocity. 
The particle position and velocity are determined by the stochastic differential equation
\begin{equation}
\dd s = 
\begin{bmatrix}
s_2 \dd t\\ 
    \left(-\gamma s_2 + u\right) \dd t + \epsilon \dd W
    \end{bmatrix}
    \end{equation}
where $\gamma=0.5$ determines the friction experienced by the particle, $u$ is the control exerted by the agent and $W$ is Brownian noise scaled by $\epsilon=0.1$. 
This system is also known as a stochastic double integrator (SDI)~\cite{Rao2001Naive}.


\quad To implement a (noisy, analog) neuromorphic agent that controls the particle dynamics, we consider a stochastic continuous-time recurrent neural network (CTRNN)~\cite{Beer1995Dynamics}. 
Let $\alpha$ denote the state variables of the CTRNN that implement the inference process, mapping observations to controls through dynamics 
\begin{equation}
\dd \alpha = \tau^{-1} \circ \left(-\alpha + A\sigma(\alpha + b) + B y\right) \dd t + \kappa \dd W
\label{eq:CTRNN}
\end{equation}
with time constants $\tau = (\tau_1,\ldots,\tau_k)$, Hadamard product $\circ$, sigmoid function $\sigma(\cdot)$,  bias vector $b=(b_1,\ldots,b_k)$, observations $y$, noise scale $\kappa$ and $k$-dimensional Brownian noise $W$. 
We further assume that the control is given by a saturating nonlinear readout of the agent state $u = \tanh(C \alpha)$ and define $\theta = (\tau, b, A,B, C)$ as the model parameters. CTRNN parameters are all initialized to zero, except the time constants, which are set to one. The number of neurons is set to $k=2$. 
The initial state $a_0 = 0$ and noise scale $\kappa = 0.01$ are fixed hyper-parameters.
For simplicity, we assume 
complete observability of the particle position and velocity. That is, the agent's observations are given by $y = s$. 
Finally, we assume that the agent's objective function is given by Eq.~\eqref{eq:objective} without discounting ($\lambda=0$) and instantaneous quadratic reward function 
\begin{equation}
    r(t) = -0.9 s_1(t)^2 - 0.1 u(t)^2
    \label{eq:objective1}
\end{equation}
penalizing large magnitudes of the particle position $s_1$ while attenuating the control amplitude $u$. 

\quad The coupled CTRNN-SDI system was implemented using Diffrax~\cite{Kidger2022Neural}; a JAX-based numerical integration package. Numerical integration was performed using a stochastic Runge-Kutta solver with a step size of $\Delta t = 0.1$~\cite{Foster2023High}. Given randomly initialized parameters, the coupled dynamics between the agent and its environment evolve in a non-teleological manner. The question is how to adapt the system in order to achieve more effective control.

\end{tcolorbox}

We assume that $x(t)$ describes the joint dynamics of the agent and the environment. We can make this more explicit by defining $x(t) = (z(t), s(t))$ with $z(t) \in \mathbb{R}^{d_z}$ the state of the agent and $s(t) \in \mathbb{R}^{d_s}$ the state of the environment. Define $f(x) = (f_z(x), f_s(x))$ and $v = (v_z, v_s)$. We may now  write the joint dynamics $\dd x = f(x) \dd v$ in terms of two coupled dynamical systems 
\begin{subequations}
\begin{align}
\dd z &= f_z(x) \dd v_z \\
\dd s &= f_s(x) \dd v_s 
\end{align}
\end{subequations}
with initial state $x_0 = (z_0, s_0)$.
Using this notation, given driving paths $v_z$ and $v_s$, the agent and the environment are completely specified by tuples $a = (f_z, z_0)$ and $e = (f_s, s_0)$, respectively.
The objective of neuromorphic computing can now be equated with defining an agent $a = (f_z, z_0)$ such that the joint dynamics maximize \eqref{eq:intelligence} as the result of sequential decision making under uncertainty by the agent, translating observations $y = g_z(s)$ into controls $u = g_s(z)$. Determining the optimal control from observational data is particularly challenging given the epistemic uncertainty of the agent about (the state of) the environment since it can only partially observe the environmental state through its senses.
Figure~\ref{fig:marr-neuromorphic}b depicts the interaction between the agent and the environment across time. 
In Box~1, we consider a concrete working example of such a coupled system, where the aim is to control the position of a stochastic particle.
\footnote{Code to reproduce all simulations in this paper is available at \url{https://github.com/marcelvangerven/neuromorphic_intelligence}.}

\subsection{The implementational level}

The implementational level of analysis asks how an intelligent system is physically realized. The neuromorphic computing paradigm assumes that there is no distinction between the physical substrate that implements an intelligent algorithm and the algorithm itself. In this sense, a neuromorphic system can be seen as a special-purpose device whose sole purpose is to generate appropriate output in response to its input. This notion of directly using the physical substrate for compute is also referred to as in-physics or in-materia computing~\cite{Nakajima2022Physical,Markovic2020Physics,Kaspar2021Rise,Jaeger2023Formal}. In other words, the substrate {\em is} the compute~\cite{Laydevant2024Hardware}. Furthermore, unlike conventional von Neumann architectures, which separate processing and memory units, leading to inefficiencies due to the Von Neumann bottleneck~\cite{Hassabis2017Neuroscienceinspired}, neuromorphic architectures integrate these units to achieve in-memory computing, removing the distinction between memory and compute. 

Under the dynamical systems approach, we achieve this equivalence between an algorithm and its physical realization by assuming that the neuromorphic agent's state equation $f_z$ implements the equations of motion of the corresponding physical system. Given the joint initial state $x_0$, the physical system is assumed to adapt towards more desirable states as it interacts with its environment as dictated by the driving paths $v$.  
This mapping is multiply realizable using various physical substrates, ranging from digital and analog silicon-based technologies~\cite{Horowitz2014Computings,Chen2020Classification} to emerging materials such as memristors~\cite{Pedretti2021InMemory}, phase-change memory~\cite{Wong2010Phase}, spintronic devices~\cite{Bhatti2017Spintronics,Grollier2020Neuromorphic,Zink2022Review}, photonic systems~\cite{Brunner2025Roadmap}, electrochemical systems~\cite{VanDeBurgt2017Nonvolatile,Stuhlmuller2025Neuromorphic,Han2022Iontronics} and biological substrates~\cite{Smirnova2024Biocomputing,Kagan2022Vitro}. In terms, of efficiency, these unconventional computing paradigms offer orders of magnitude better energy efficiency compared to conventional digital hardware, up to the Landauer limit~\cite{Landauer1961Irreversibility} and beyond~\cite{Bennett1973Logical,Fredkin1982Conservative}. 

Even though neuromorphic algorithms should ultimately run in neuromorphic substrates,  conventional hardware continues to fulfill an important role since it acts as a testbed for simulating neuromorphic agents via numerical integration of the underlying differential equations. Field-programmable gate arrays (FPGAs) are of particular interest here since this reconfigurable digital hardware combines fidelity and flexibility to pave the way for high-throughput simulation of neuromorphic systems~\cite{Farsa2025Reconfigurable,Karamimanesh2025Spiking}.

\section{Experience-dependent learning}
\label{sec:learning}

The key question remains how an agent can modify its behavior such as to maximize the expected return. One way to do so, is by learning from experience, that is, by adapting its dynamics based on incoming observations such as to reach more desirable states. 

\subsection{Automatic differentiation}

The conventional way to train deep learning models in dynamic control settings is by performing policy gradient ascent on the model parameters $\theta$ via backpropagation through time (BPTT)~\cite{Werbos1990Backpropagation}. In case of parameterized dynamical systems, BPTT amounts to reverse-mode automatic differentiation on the unrolled computational graph that is defined by the state equations and the numerical solver~\cite{Baydin2018Automatic,Kidger2022Neural}. This involves iteratively running a forward pass to simulate the coupled system followed by a backward pass to transmit gradient information for credit assignment. 
However, the backpropagation algorithm has several issues~\cite{Crick1989Recent}. First, it is non-parsimonious as it requires additional mechanisms to implement the backward pass. Second, it interleaves forward and backward passes, meaning that it continuously needs to switch between inference and learning. This is also referred to as update locking. Third, the backward pass requires that presynaptic weights have access to postsynaptic weights, making the algorithm non-local. This is referred to as the weight transport problem. Fourth, it requires transmission of exact gradients, which is generally impossible in physical systems that are subject to intrinsic noise. Finally, and perhaps most importantly, BPTT is impossible to implement in a physically realistic manner exactly because it moves against the arrow of time, preventing online learning.
An alternative approach is provided by real-time recurrent learning (RTRL), which is equivalent to running forward-mode automatic differentiation~\cite{Williams1989Learning}. While this solves the issues associated with running a backward pass, it is less efficient than BPTT and still relies on the storage and transmission of exact gradients.

\subsection{Noise-based learning}
In contrast to the conventional approach, the dynamical systems approach assumes that learning emerges purely from simulating the agent's equations of motion using operations that are local in space and time. That is, we focus on online, in-situ learning instead of the more conventional off-line, ex-situ approach to learning as implemented in conventional AI systems using the backpropagation algorithm~\cite{Werbos1974Regression}. Such lifelong,  continual learning~\cite{Kudithipudi2022Biological}
 can be realized by incorporating the learning mechanism in the system dynamics.
To this end, let us partition the agent state as $z = (\alpha, \phi)$, where we distinguish variables $\alpha$ whose dynamics implement inference, as before, from variables $\phi$, whose dynamics implement learning, 
That is, inference refers to the reasoning process which translates observation into action, whereas learning refers to changes in the reasoning process based on experience-driven adaptation. 
The agent dynamics are described by
\begin{subequations}
\begin{align}
\dd{\alpha} 
&= f_\alpha(x) \dd{v}_\alpha \label{eq:sde_inference}\\
\dd{\phi} 
&= f_\phi(x) \dd{v}_\phi  
\label{eq:sde_learning} 
\end{align}
\end{subequations}
where the dynamics of $\phi$ are assumed to induce increasingly optimal states $a$. 

The challenge is to define  $f_z = (f_\alpha, f_\phi)$ and $z_0 = (\alpha_0, \phi_0)$ in such a way that optimal behavior emerges, purely from forward simulating the equations of motion of the agent-environment system. From a dynamical systems perspective there is no difference between inference and learning, other than that the latter is assumed to evolve at a slower time scale. This is also known as a fast-slow system~\cite{Kuehn2015Multiple}. For instance, in case of biological agents, the fast system would determine neuronal dynamics whereas the slow system describes structural or functional changes due to neuronal plasticity. 
The question remains how an agent can modify its parameters to learn from experience purely by following its equations of motion.



An appealing approach to online learning is provided by stochastic perturbation methods; derivative-free zero-order optimization methods that approximate the gradient by evaluating the objective function at different points to find an optimum~\cite{Conn2009Introduction}. 
These methods exploit process noise that affects system dynamics as a way to drive parameter values in more optimal directions, implementing credit assignment via stochastic approximations of the gradient~\cite{Barto1983Neuronlike,Cauwenberghs1992Fast,Dalm2023Effective,Fernandez2024OrnsteinUhlenbeck}. Stochastic  perturbation methods view noise as a feature rather than a bug~\cite{Maass2014Noise}, making them ideally suited for implementation in (noisy, analog) neuromorphic systems~\cite{Koenders2025Noisebased}. This principle is also well-aligned with ideas about the functional role of noise in biological systems~\cite{Burnod1989Consequences,Piwkowska2008Characterizing,Deco2009Stochastic,Shimizu2021Computational,Kappel2015Network}.

\begin{tcolorbox}[float=t, title=Box 2: Learning to control a stochastic particle, colback=white, colframe=blue, colbacktitle=lightgray, coltitle=black, fonttitle=\bfseries]
\label{box:box2}

\vspace{-5pt}
\begin{center}
\includegraphics[width=\textwidth]{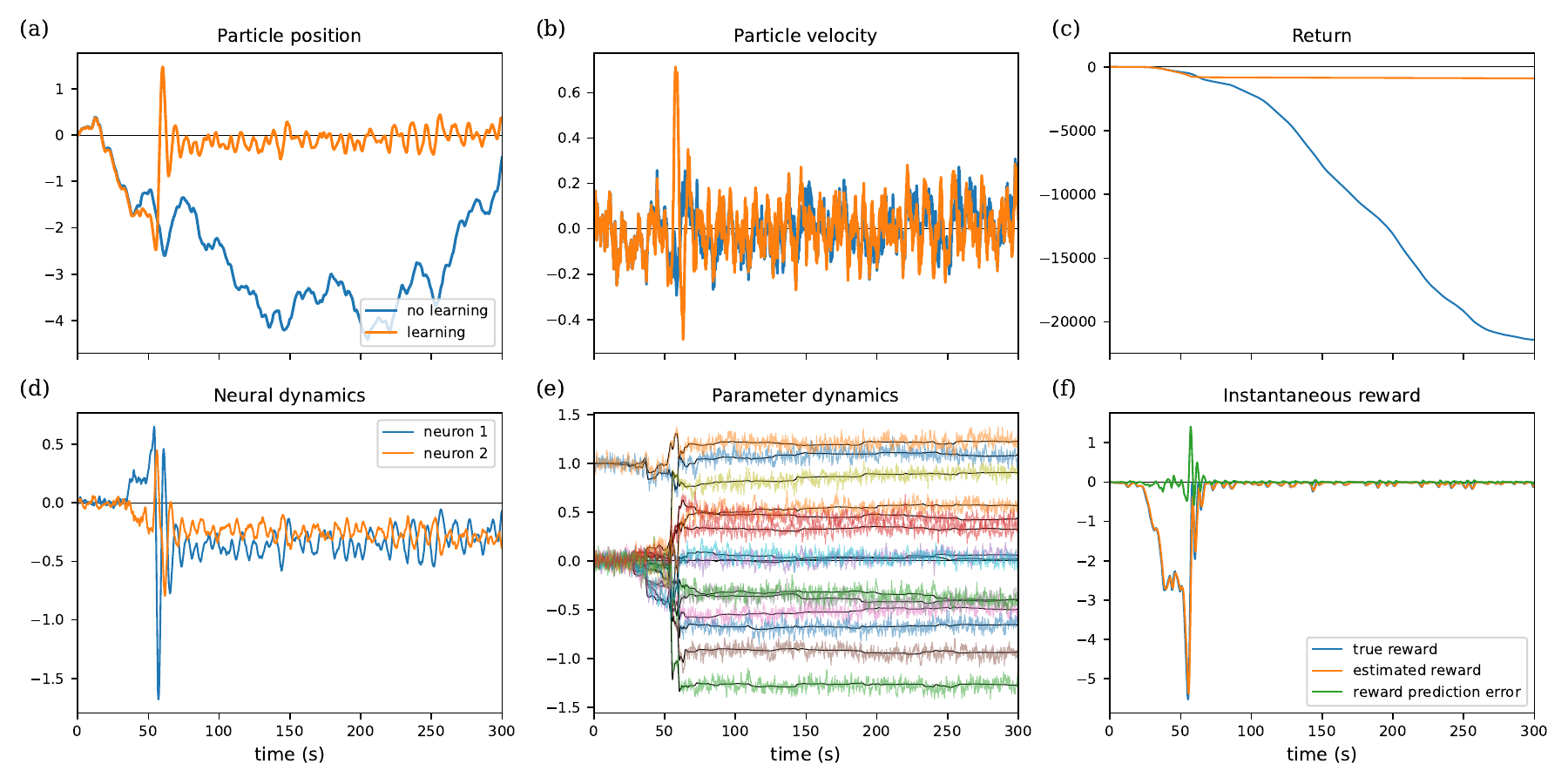}
\end{center}
\vspace{-10pt}

To demonstrate the use of noise as a mechanism for learning, we applied OUA to our working example. To this end, we extend the CTRNN state to $z = (\alpha, \phi)$ where the dynamics of $\alpha$ are given by \eqref{eq:CTRNN} and $\phi = (\theta, \mu, \nu)$, where $\theta$ are the CTRNN parameters and $\nu$ is computed from the reward function \eqref{eq:objective1}.

\quad  We compare SDI dynamics with and without learning. Learning is induced via Ornstein-Uhlenbeck adaptation (OUA). OUA hyper-parameters given by $\lambda = 2$, $\sigma=0.1$, $\eta = 5$ and $\rho=2$. Panels a-c show the particle position, particle velocity and return (cumulative reward) with and without learning. As shown in the figure, without learning, the particle's position fluctuates about since the dynamics are effectively uncontrolled in this case, which is associated with a low return (Panels a-c, blue curves). 
In contrast, with learning, the dynamics adapt such as to maintain the particle position around zero (Panels a-c, orange curve).
Panels d-f show the agent dynamics as it learns to control the particle. Panel d shows the evolving neuronal state $\alpha$, Panel e shows changes in the parameters $\theta$, and Panel f depicts the true reward $r$, estimated reward $\nu$ and reward prediction error $\delta$ during OUA learning. 

\quad Hence, for this problem, the proposed OUA learning mechanism indeed modifies the CTRNN parameters in such a way that a more desirable state is obtained, purely by simulating the equations of motion of the system at hand. 

\end{tcolorbox}

As an example of noise-based learning, let us consider Ornstein-Uhlenbeck adaptation (OUA) -- a recently proposed mechanism that incorporates noise-based learning in the agent's state equations~\cite{Fernandez2024OrnsteinUhlenbeck}. 
Crucially, in OUA, learning emerges from evaluating the equations of motion of the augmented system.
It induces parameter changes by introducing state variables $\phi = (\theta, \mu, \nu)$ that implement learning. Here, $\theta \in \mathbb{R}^n$ are the parameters on which the dynamics of inference variables $z$ depend, $\mu \in \mathbb{R}^n$ is a vector of mean parameter values and $\nu \in \mathbb{R}$ is a reward estimate. 
For each parameter $\theta_i$, the dynamics are defined by an Ornstein-Uhlenbeck (OU) process
$
\dd \theta_i(t) = \lambda (\mu_i(t) - \theta_i(t)) \dd t + \sigma \dd W(t)
$, 
where $\lambda$ is a rate parameter and $\sigma$ determines how quickly parameters $\theta_i$ diffuse away from their mean $\mu_i$ due to Brownian noise $W$. Starting at $\mu_0 = \theta_0$, the OUA dynamics implement an exploration process about the current parameter mean. Learning is induced via the parameter mean $\mu \in \mathbb{R}^n$ through dynamics
$
\dd \mu_i(t) = \eta \delta(t) (\theta_i(t) - \mu_i(t)) \dd t
$, 
where $\eta$ is a rate parameter and $\delta(t) = J(t) - \nu(t)$ is the reward prediction error (RPE), acting as a learning signal for credit assignment. The RPE quantifies the difference between the expected return under the parameters $\theta$ and the expected return under the mean $\mu$. 
Following~\cite{Fernandez2024OrnsteinUhlenbeck}, we ignore future rewards and use $\delta(t) = r(t) - \nu(t)$, where $r(t)$ is the observed reward and $\nu(t)$ is an average reward estimate whose dynamics are given by
$\dd \nu(t) = \rho (r(t) - \nu(t)) \dd t 
$
with rate parameter $\rho$ and initial state $\nu(0) = 0$. 

Box~2 demonstrates that OUA as an example of noise-based learning can indeed learn to effectively control the stochastic particle introduced in Box~1.
Note further that the same OUA mechanism can be applied to the model hyper-parameters. For example, as shown in~\cite{Fernandez2024OrnsteinUhlenbeck}, dynamic updating of the noise variance $\sigma$ allows for adaptively increasing and decreasing the noise variance, effectively trading off between exploration and exploitation. 
This exploration-exploitation dilemma is a key problem faced by any intelligent agent faced with epistemic uncertainty~\cite{Sutton2015Reinforcement}. The need to trade off between information-seeking and reward-seeking behavior is known as dual control in the control theory community~\cite{Feldbaum1960Dual}.

\section{Differential genetic programming}
\label{sec:evolution}


The dynamical systems approach to the development of intelligent agents is concerned with setting up  equations of motion that implement learning, inference and control.  The previous section demonstrated that we can set up the equations of motion of a neuromorphic system in such a way that the system self-organizes towards increasingly optimal states.  Here, we assumed that adaptation takes place at an ontogenetic timescale, that is, during the lifetime of an intelligent agent. The question remains, however, if the equations that implement learning, inference and control are optimal themselves. In this section, we will consider how we can evolve neuromorphic agents at a phylogenetic timescale through a process of natural selection. 
From this evolutionary perspective, the goal is to evolve populations of agents 
$$
\mathcal{A}_{0} \rightarrow \mathcal{A}_1 \rightarrow \cdots \rightarrow \mathcal{A}_N
$$
with $\mathcal{A}_i$ the agent population at the $i$-th generation such that the average fitness ${F}_i = \frac{1}{||\mathcal{A}_i||} \sum_{a \in \mathcal{A}_i} J(a)$ of the population increases across generations, where an individual agent's fitness is equated with $J(a)$. 

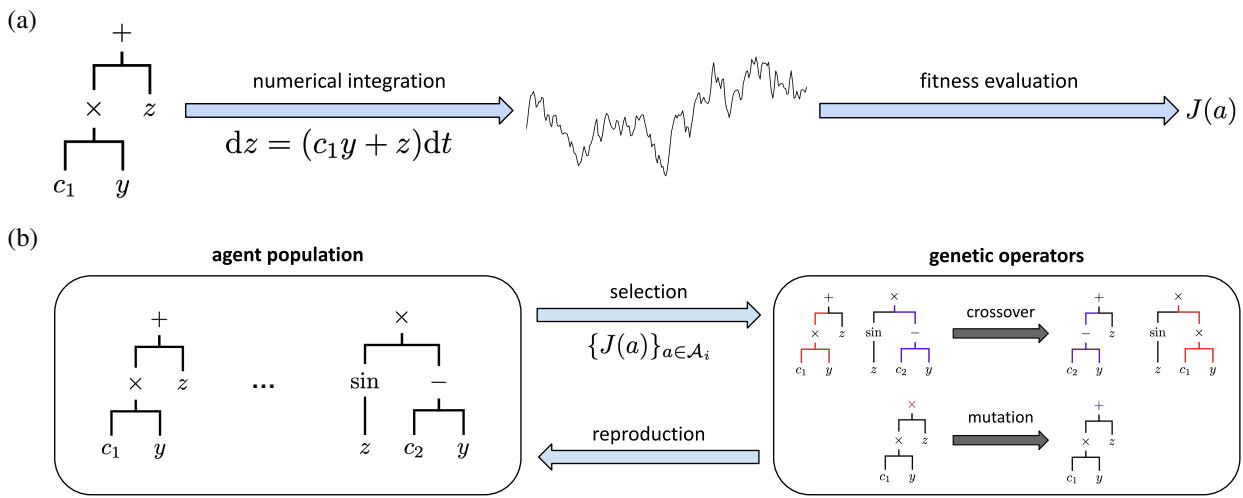
\begin{figure}[!ht]
  \centering
  
  \begin{subfigure}{0.90\textwidth}
    \centering
    \setlength{\unitlength}{1cm}
    \begin{picture}(0,0)
      \put(-8.5,0){\makebox(0,0)[l]{(a)}} 
    \end{picture}
    \includegraphics[width=\linewidth]{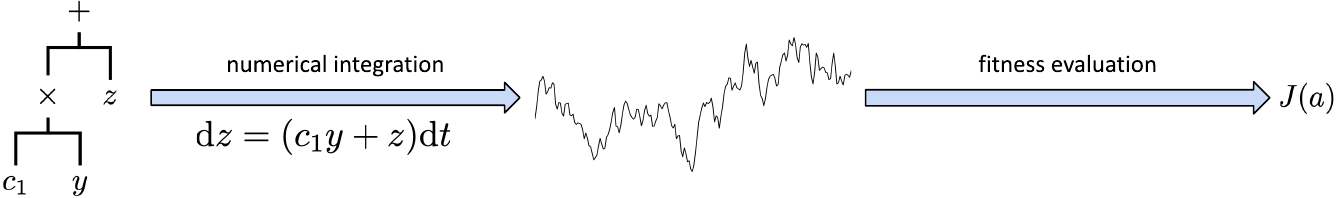}
  \end{subfigure}

  \vspace{0.5cm}
  
  \begin{subfigure}{0.90\textwidth}
    \centering
    \setlength{\unitlength}{1cm}
    \begin{picture}(0,0)
      \put(-8.5,0){\makebox(0,0)[l]{(b)}} 
    \end{picture}
    \includegraphics[width=\linewidth]{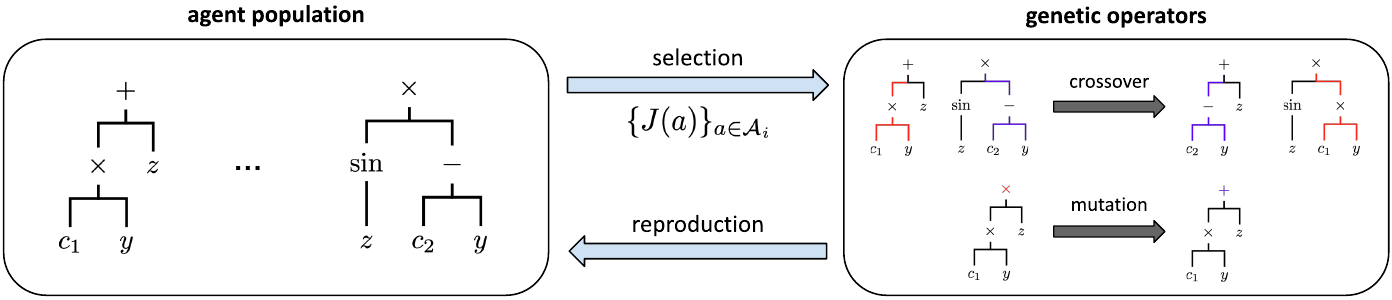}
  \end{subfigure}
  
  \caption{Evolving neuromorphic agents using DGP. a) The state equation describing a neuromorphic agent can be represented by a computational graph consisting of function nodes and leaf nodes. The agent-environment interaction is simulated using numerical integration, based on which the agent's fitness can be computed. We here consider a scalar ODE. This can be  extended to more complex systems of differential equations whose underlying equations are represented by multiple trees. 
  b) 
    Agents are preferentially selected for reproduction based on their fitness. The new population is evolved using crossover and mutation operations.
    }
\label{fig:gp}
\end{figure}

Evolving neuromorphic systems constitutes a truly bottom-up approach for the development of neuromorphic intelligence.
Evolutionary adaptation affords changes towards optimality that cannot be learned through experience-dependent learning. This particularly pertains to the non-differentiable symbolic structure of the state equation as well as those changes whose optimality cannot be sensed locally in space and time. This pertains, e.g., to evolving the structure of sensory and motor components that link the agent to its environment, evolving objective functions that can be locally estimated and act as surrogates to the true (but possibly unknowable) objective $J$, or evolving the slow part of the system to induce more effective learning.

\begin{tcolorbox}[float=t, title=Box 3: Evolving control of a stochastic particle, colback=white, colframe=blue, colbacktitle=lightgray, coltitle=black, fonttitle=\bfseries]
\label{box:box3}

 \begin{center}
     
    \setlength{\unitlength}{1cm}
    \vspace{5pt}
    \includegraphics[width=\linewidth]{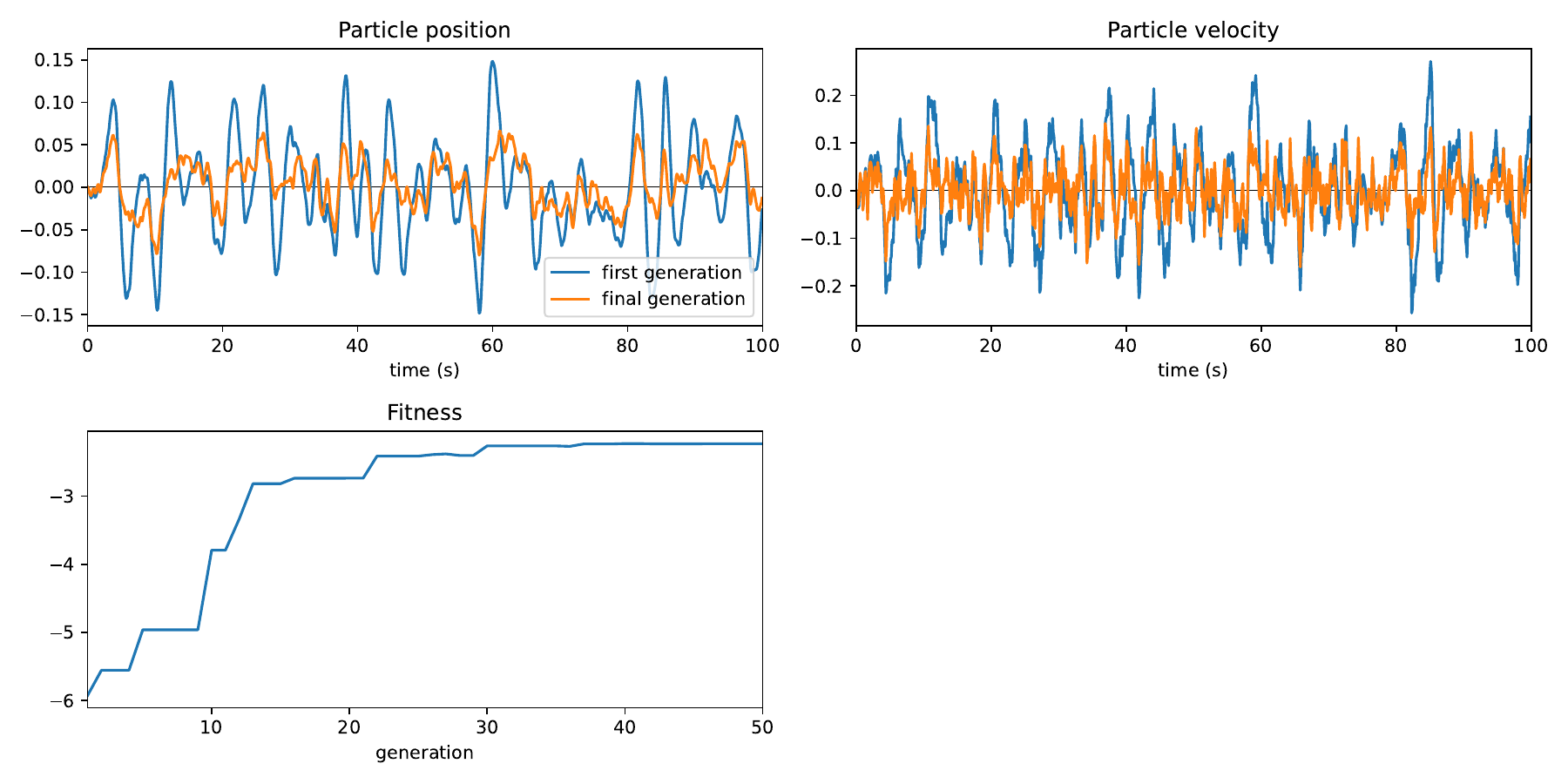}
    
    \begin{picture}(0,0)
      \put(-8.25,8.85)
      {\makebox(0,-0.7)[l]{(a)}} 
      \put(0.05,8.85){\makebox(0,-0.7)[l]{(b)}} 
      \put(-8.25,4.75){\makebox(0,-0.7)[l]{(c)}}
      \put(0.05,4.75){\makebox(0,-0.7)[l]{(d)}}
      
      
     \put(-0.05,1.47){
        \raisebox{\height}{\scalebox{0.95}{
    \begin{forest}
for tree={
  edge path={
    \noexpand\path [\forestoption{edge}]
      (!u.parent anchor) -- +(0,-5pt) -| (.child anchor)\forestoption{edge label};
  },
  edge={line width=1pt},
  parent anchor=south,
  child anchor=north
}
[$\times$
          [\small{$-1.15$}]
          [$y_1$]
]
\end{forest}
}}}

 \put(1.5,-0.4){
 \raisebox{\height}{
        \scalebox{0.95}{
    \begin{forest}
for tree={
  edge path={
    \noexpand\path [\forestoption{edge}]
      (!u.parent anchor) -- +(0,-5pt) -| (.child anchor)\forestoption{edge label};
  },
  edge={line width=1pt},
  parent anchor=south,
  child anchor=north
}
[$-$
      [$\times$
          [\small{$-6.14$}]
          [$z_2$]
      ]
      [$\times$
          [\small{$2.07$}]
          [$y_1$]
      ]
  ]
\end{forest}
}}}

 \put(5.3,-0.4){
 \raisebox{\height}{
        \scalebox{0.95}{
    \begin{forest}
for tree={
  edge path={
    \noexpand\path [\forestoption{edge}]
      (!u.parent anchor) -- +(0,-5pt) -| (.child anchor)\forestoption{edge label};
  },
  edge={line width=1pt},
  parent anchor=south,
  child anchor=north
}
[$+$
      [$\times$
          [\small{$2$}]
          [$z_1$]
      ]
      [$\times$
          [\small{$6$}]
          [$z_2$]
      ]
  ]
\end{forest}
}}}

    \end{picture}
 \end{center}

  \vspace{-25pt}

To enable control of a stochastic particle, 
rather than specifying the agent as a CTRNN, we evolve the symbolic equations that together define the agent's dynamics. We define two internal states $z = (z_1, z_2)$ for the agent and use observations $y = (y_1, y_2)$ denoting the particle's position and velocity. Function nodes were chosen to be elements in $\{\times, +, -\}$, providing a complete basis for polynomial functions. To implement DGP for the SDI, we made use of Kozax -- a flexible and efficient genetic programming framework developed in JAX~\cite{DeVries2025Kozax}. DGP was run for 50 generations based on ten interacting populations with a population size of 100. 
Panels~a and b show the particle position and velocity, respectively, for the best agent in the first and last generation when evolving the control of a stochastic particle.  
Panel~c shows increasing maximum fitness of the evolving agent population, demonstrating that symbolic expressions can be evolved to achieve more effective control. Panel~d shows the best symbolic expression found in the last generation. Here, the first tree represents the state equation $f_1(z, y)$ for variable $z_1$, the second tree represents the state equation $f_2(z, y)$ for variable $z_2$ and the third tree represents the readout $u = g(z)$.
\end{tcolorbox}

To emulate evolutionary adaptation, we may resort to genetic programming (GP)~\cite{Cramer1985Representation,Koza1994Genetic}. 
Genetic programming is an evolutionary algorithm for learning computer programs. GP maintains a population of individuals that each embody a computer program represented in terms of a computational graph. 
This graph is constructed from predefined sets of function nodes and leaf nodes. Function nodes cover mathematical (unary and binary) functions, while leaf nodes describe variables and constants (nullary functions). GP seeks to evolve populations with increasing fitness through crossover and mutation operations. To maintain diversity, prevent premature convergence and enable specialization, typically, several subpopulations are maintained with different crossover and mutation rates.

In the dynamical systems approach to AI, we may employ GP to evolve agents $a = (f_z, z_0)$. The initial state $z_0$ simply requires evolving a state vector. For ease of exposition, we will assume that the initial state is fixed to $z_0 = 0$. In contrast, we interpret $f_z$ as a computer program $x \mapsto \dd z$ that takes input $x$ and generates differentials $\dd z$.  I refer to the genetic programming approach to evolving dynamical systems as {\em differential genetic programming} (DGP). 
Figure~\ref{fig:gp} illustrates the use of DGP to evolve neuromorphic agents. Figure~\ref{fig:gp}a shows how the state equation of a scalar ordinary differential equation, representing the neuromorphic agent, can be represented as a parse tree. The fitness of this state equation can be evaluated by computing the return as the agent-environment interaction is simulated. Figure~\ref{fig:gp}b shows how differential genetic programming evolves populations of agents through selective crossover and mutation. The generalization to more complex dynamical systems consisting of multiple state variables, state equations representing both drift and diffusion and/or readout mechanisms that translate states to control output is straightforward, requiring the evolution of a multitree (set of parse trees). 
De Vries et al.~\cite{DeVries2024Discovering} have shown that the state equations of dynamical systems can be evolved to effectively control classical control problems. 
To demonstrate evolving populations of agents with increasing average fitness ${F}$, we again consider our working example of controlling a stochastic particle in Box~3.

\section{Discussion}

In this paper we have seen how neuromorphic agents, interpreted as a system of differential equations, implementable in hardware,  can be effectively learnt, either through a process of experience-dependent learning at an ontogenetic timescale or by evolving the equations as symbolic expressions using differential genetic programming at a phylogenetic timescale. This suggests that \emph{dynamical systems are all you need}~\cite{Vaswani2017Attention,Ramsauer2021Hopfield,Feng2024Were} to develop more efficient and effective neuromorphic AI systems. 
A key assumption here is that intelligent adaptive behavior must be an emergent phenomenon which arises exclusively from following the physical equations of motion that describe the neuromorphic agent. 
In the following, we discuss this approach in more detail and point out directions for future research.

We specifically considered Ornstein-Uhlenbeck adaptation as a mechanism for learning that is driven by the intrinsic noise inherent to analog systems. This nicely complements recent work, which showed that the self-organizing behavior of neuromorphic nanowire networks can be represented as an OU process~\cite{Milano2025Selforganizing}. While OUA provides a proof of principle of noise-based learning, several limitations and open questions remain. OUA assumes that the reward prediction error $\delta$ -- as the difference between the current reward and the average reward -- is a suitable signal for credit assignment. Whereas this was shown to be effective in low-dimensional linear systems such as the stochastic double integrator, alternative formulations of noise-based learning are likely required when dealing with large-scale nonlinear control problems. Note further that OUA -- which can be seen as a continuous-time local formulation of weight perturbation~\cite{Jabri1991Weight,Cauwenberghs1992Fast,Werfel2005Learning} -- is but one proposed mechanism for the development of physical learning machines. For example, equilibruim propagation is an alternative approach, which proceeds by comparing the steady state of a system run in a free phase and weakly clamped phase and using this as a learning signal~\cite{Scellier2017Equilibrium,Stern2021Supervised}. 

We further committed ourselves to using continuous-time recurrent neural networks of the form~\eqref{eq:CTRNN} when specifying the inference mechanism of the neuromorphic agent. Other systems of equations defining liquid time-constant models~\cite{Hasani2021Liquid} or coupled oscillator models~\cite{Effenberger2025Functional} may turn out to be more suitable in terms of parameter efficiency as well as physical realizability. Furthermore, we exclusively considered analog systems described by continuous-valued state variables. In practice, we may want to embrace mixed analog-digital systems that, like our own brains, communicate via spikes~\cite{Izhikevich2018Dynamical,Abbott2016Building}. Such systems can be more energetically efficient given their event-driven nature. 

Concerning physical realizability, it is assumed that the system of equations specifying the agent can be mapped to specific neuromorphic hardware as an equivalent circuit. Whether or not this is the case depends on if such a one-to-one mapping exists. One step in this direction is the demonstration that physical (telegraph) noise generated by stochastic magnetic tunnel junctions can be used to drive learning~\cite{Koenders2025Noisebased}. 
Note further that, even if such a mapping exists, results obtained in simulations may not carry over to physical systems due to the limited precision of numerical integration. 
To enforce algebraic constraints, such as respecting Kirchoff's laws, specialized solvers may be needed such as algebraic differential equations of the form $M(x) \dd x = f(x) \dd v$, where the mass matrix $M \colon \mathbb{R}^d \to \mathbb{R}^{d \times d}$ ensures that the dynamics remain on the constraint manifold~\cite{Romisch2003Stochastic}. 
Next to this, unmodelled circuit properties such as parasitic currents may also cause misalignment between the (SPICE) simulation and the physical realization of the neuromorphic circuit at hand.

To evolve neuromorphic intelligence, 
the concept of differential genetic programming was introduced. Here, the idea is that neuromorphic circuits can be evolved using principles developed in evolutionary computing~\cite{Koza1999Design}. Using our working example, it was shown that it is indeed possible to evolve the state equation of a neuromorphic agent such that it learns to compensate for stochastic perturbations. Moreover, representation of the state equation in terms of a symbolic expression facilitates interpretation of how the agent achieves its goal, contributing to the development of more transparent and explainable AI systems. 
Note however that to date is remains notoriously hard to evolve  computational graphs that effectively solve large-scale problems. For example, GP is know to suffer from code bloat, referring to the uncontrolled growth of computational graphs without corresponding increases in fitness~\cite{Koza1998Genetic,Banzhaf1998Genetic,Langdon2013Foundations}.

If these problems can be overcome, genetic programming may become the ultimate approach in artificial intelligence as it affords truly end-to-end learning of intelligent systems. It may also become feasible to evolve the learning mechanism itself. That is, by pitting neuromorphic agents against unpredictable environments, learning mechanisms competitive with backpropagation might emerge as a strategy for the agent to adapt during its lifetime~\cite{Zador2019Critique}. This is also referred to as meta-learning or learning to learn~\cite{Bengio1992Learning,Hospedales2020MetaLearning,Jordan2021Evolving}. 
Here we envision that zero-order gradient methods that embrace noise as a mechanism for learning may turn out to be essential. Finally, emergent adaptive mechanisms may not only pertain to experience-dependent changes but also developmental changes that unfold during the agent's lifetime~\cite{Haber2018Emergence}, as well as morphological changes that modify the coupling between an agent and its environment~\cite{Arbel2025Mechanical}. 

Let us return to the concept of neuromorphic computing. Given the large number of involved disciplines it is sometimes said that there exist as many definitions of neuromorphic computing as there are neuromorphic researchers.   This paper proposes to define neuromorphic computing as {\em the capacity of a dynamical system to realize useful computations exclusively through the intrinsic evolution prescribed by its governing equations of motion}. This is in contrast to conventional digital computing, which relies on explicit symbolic manipulation and externally imposed control structures.
Given its multiple realizability, neuromorphic computing should not be defined by its physical substrate. Nonetheless, the human brain offers a paradigmatic example of such a system, providing the primary source of inspiration for neuromorphic design.

Whereas neuromorphic computing may relate to any form of computation that is useful from the perspective of the user, neuromorphic {\em intelligence} aims to create systems that are as efficient and capable as the human brain. This paper argues that neuromorphic intelligence can be conveniently described in terms of Marr's levels of analysis, linking intelligence -- as the ability of agent to adapt to changing circumstances -- to algorithms defined as systems of differential equations that can be mapped to physical substrates.
This provides a roadmap for the neuromorphic community, providing a bridge between multiple disciplines that embody different levels of the neuromorphic value chain, ranging from theory to applications. Which applications we will pursue critically depends on ourselves;   
especially if we want to live in a society that is based on human values other than buying or selling~\cite{Wiener1948Cybernetics,Rickover1965Humanistic}.

\section*{Acknowledgements}

This publication is part of the project Dutch Brain Interface Initiative (DBI2) with project number 024.005.022 of the research programme Gravitation, which is financed by the Dutch Ministry of Education, Culture and Science
(OCW) via the Dutch Research Council (NWO). 
The author thanks Sigur de Vries for his helpful comments.






\bibliographystyle{naturemag-doi}
\bibliography{main}


\end{document}